\documentclass{article}
\usepackage{graphicx} 
\usepackage{amsmath}
\usepackage{amsfonts}
\usepackage{latexsym}
\usepackage{amssymb}
\usepackage{amsmath}

\usepackage{times}
\normalfont
\usepackage[T1]{fontenc}
\usepackage{graphicx}

\newcounter{thetheorem} 
\newtheorem{theorem}[thetheorem]{\bf Theorem}
\newcounter{theproposition} 

\newcounter{thelemma} 

\newcounter{definition}
\newenvironment{definition}[1][]{\refstepcounter{definition}\noindent%
   \textbf{Definition~\thedefinition. #1} \rmfamily}{\medskip}

\title{Cogent argument extensions are weakly admissible but not vice versa}
\author{Gustavo Bodanza\footnote{Departamento de Humanidades, Universidad Nacional del Sur and Instituto de Investigaciones Econ\'{o}micas y Sociales del Sur (IIESS-UNS/CONICET), Argentina.}}

\date{}

\begin{document}

\maketitle

\section{Introduction}

In this research note, we show the relationship between two non-admissible argumentation framework semantics: cogent  \cite{bodanzatohme,bts} and weakly admissible semantics \cite{baumann20}. We prove that, while cogent extensions are weakly admissible, the converse is not true.

\section{Argumentation frameworks}

Dung's argumentation frameworks \cite{dung} are intended to represent debates or argument situations in which arguments interact through attacks.\\

\begin{definition}
    \cite{dung} An \emph{argumentation framework} is a structure $F$ $=$ $(A, R)$, where $A$ is a set whose elements are called \emph{arguments} and $R\subseteq A\times A$ is a relation of \emph{attack} among arguments.
\end{definition}

We say that $x$ \emph{attacks} $y$ iff $(x, y)\in R$. Moreover, given susbets $S, S'\subseteq A$ we will also say that $S$ attacks $x$ (resp., $x$ attacks $S$) iff $y$ attacks $x$ (resp., $x$ attacks $y$) for some $y\in S$, and $S$ attacks $S'$ iff there exists $x$ in $S$ such that $x$ attacks $S'$.

 The arguments emerging as victorious are collected in subsets called \emph{extensions}, which satisfy certain specified ``semantics'' or requirements for defense. All of Dung's semantics satisfy the following ``admissibility'' criterion.\\

\begin{definition}
    A subset $S\subseteq A$ is \emph{admissible} iff:
    \begin{enumerate}
        \item for all $x, y\in S$, $(x, y)\notin R$ ($S$ is conflict-free);
        \item for all $x\in A$, if $x$ attacks $S$ then $S$ attacks $x$. ($S$ is defensible)
        \end{enumerate}
\end{definition}

\section{Non-admissible semantics: cogency and weak admissibility}

Several authors have introduced argumentation framework semantics that depart from the notion of admissibility \cite{ver,bg05,bodanzatohme,bts,baumann20}. The aim is to capture other, weaker yet sensible defense criteria. Here we deal with cogency and weak admissibility criteria. Both are intended to avoid the undesired effects of self-attacking arguments. The idea is that any attacks resulting from those arguments should be ignored because they are irrational. For example, given $F$ $=$ $(\{a, b\}, \{(a, a), (a, b)\})$, we have that $\emptyset$ is the only admissible subset of arguments, but $b$ should be accepted because it is only challenged by the self-attacking argument, $a$.\\

In \cite{bodanzatohme, bts}, the notion of \emph{cogency} is proposed to capture the intended semantics.

\begin{definition} \cite{bts} Given an argumentation framework $F$ $=$ $(A, R)$, and two subsets $E$, $E'$ $\subseteq$ $A$, we say that $E$ is \emph{at least as cogent as} $E'$, in symbols, $E\geq_{cog}E'$, iff $E$ is admissible in the restricted argumentation framework $F\mid_{E\cup E'}=(A, R\mid_{E\cup
E'})$. We say that $E$ is \emph{strictly more cogent than}
$E'$, in symbols, $E>_{cog}E'$, iff $E\geq_{cog}E'$ and not $E'\geq_{cog}E$.
\end{definition}

\begin{definition} \cite{bts} Given an argumentation framework $F=(A, R)$, we say that a subset
$E\subseteq A$ is (simply) \emph{cogent} iff $E$ is maximal w.r.t. $>_{cog}$, i.e., $\forall E'\subseteq A$ $E'\not>_{cog}E$. We also define $Cog(AF)=\{E\subseteq A: E$ is
cogent$\}$.
\end{definition}

Later, in \cite{baumann20} the notion of \emph{weak admissibility} is introduced, yielding a similar semantics but with a different formal apparatus.

\begin{definition}
\cite{baumann20} Given an argumentation framework $F=(A, R)$ and $E \subseteq A$, define:
    \begin{itemize} 
    \item $E^{+}=\{a\in A: E\ \text{attacks}\ a\}$
    \item $E^{\oplus}=E\cup E^{+}$
\end{itemize}
\end{definition}

\begin{definition} \cite{baumann20} Given an argumentation framework $F=(A, R)$, let $E \subseteq A$. The $E$-reduct of $F$ is the argumentation framework $F^E=\left(E^*, R \cap\left(E^* \times E^*\right)\right)$ where $E^*=A \backslash E^{\oplus}$.
\end{definition}

\begin{definition} \cite{baumann20} Given $F=(A, R)$, $E \subseteq A$ is called \emph{weakly admissible} in $F\left(E \in a d^w(F)\right)$ iff
\begin{enumerate} 
\item  $E$ is conflict-free, and
\item for any attacker $y$ of $E$ we have $y \notin \bigcup a d^w\left(F^E\right)$.
\end{enumerate}
\end{definition}

\section{Relationship between cogency and weak admissibility}

The following results demonstrate the relationships between cogency and weak admissibility.

\begin{theorem} $Cog(F)$ $\subseteq$ $ad^{w}(F)$.
\end{theorem}

\noindent {\emph Proof.} Let $E\in Cog(F)$ and assume for the sake of contradiction that $E\not\in ad^{w}(F)$. From this last, it follows that either $E$ is not conflict-free or that there exists $E'$ $\in$ $ad^{w}(F^{E})$ such that $E'$ attacks $E$. But, because $E$ is cogent, $E$ is conflict-free. Hence, there exists $E'$ $\in$ $ad^{w}(F^{E})$ such that $E'$ attacks $E$. Now, $E'$ is conflict-free ---since it is weakly admissible in $F^{E}$--- and, by definition of $F^{E}$, $E$ does not attack $E'$. Then, we have that $E'$ is admissible in $F\mid_{E\cup E'}$. But $E$ is not, since it does not counterattack $E'$. Therefore, $E'>_{cog}E$, which contradicts the hypothesis that $E\in Cog(F)$.$\Box$\\

The converse of the previous theorem is not true in general, as can be seen from the following example. Consider $F=(\{a, b, c\},$ $\{(a, b), (b, c)\})$. Then $\{c\}$ is weakly admissible, since $b$ ---the attacker of $c$--- does not belong to any weakly admissible set in $F^{\{c\}}$. However, it is not cogent, since $\{b\}$ is admissible in $F\mid_{\{a, b\}}$ but $\{a\}$ is not. Instead, if we look at maximal (w.r.t. $\subseteq$) weakly admissible and cogent subsets, we get an agreement in $\{a, c\}$. However, this correspondence cannot be generalized to any argumentation framework either.  There could exist maximal weakly admissible subsets which are not cogent, as in the argumentation framework $(\{a, b, c, d\}, \{(a, b), (b, c), (c, a), (b, d)\})$, where $\{d\}$ is such a subset.\\

In consequence, we can say that cogent semantics is more skeptic than weakly admissible semantics or, conversely, weakly admissible semantics is more credulous than cogent semantics.\\

\section*{Acknowledgments}
Partially supported by Agencia I+D+i (PICT 2021-0075) and Universidad Nacional del Sur (PGI 24/I305), Argentina.

\end{document}